\documentclass{article}
\usepackage{ijcai23}

\usepackage{times}
\usepackage{soul}
\usepackage{url}
\usepackage[hidelinks]{hyperref}
\usepackage[utf8]{inputenc}
\usepackage[small]{caption}
\usepackage{graphicx}
\usepackage{amsmath}
\usepackage{amsthm}
\usepackage{booktabs}
\usepackage{algorithm}
\usepackage{algorithmic}
\usepackage[switch]{lineno}
\usepackage{cleveref}
\usepackage{xcolor}

\usepackage{multirow} 
\usepackage{multicol} 
\usepackage{arydshln}
\newcommand\blfootnote[1]{%
  \begingroup
  \renewcommand\thefootnote{}\footnote{#1}%
  \addtocounter{footnote}{-1}%
  \endgroup
}

\newcommand{\method}{\textsc{LEAP}}

\title{Transferring Procedural Knowledge across Commonsense Tasks
}

\author{
Yifan Jiang$^1$\and
Filip Ilievski$^{1,2}$ \and
Kaixin Ma$^3$
\affiliations
$^1$Information Sciences Institute, Viterbi School of Engineering, University of Southern California\\
$^2$Department of Computer Science, Faculty of Science, Vrije Universiteit Amsterdam\\
$^3$Tencent AI Lab, Bellevue, WA\\
\emails
\{yifjia,ilievski\}@isi.edu,
f.ilievski@vu.nl,
kaixinma@global.tencent.com
}
\begin{document}

\maketitle

\begin{abstract}
Stories about everyday situations are an essential part of human communication, motivating the need to develop AI agents that can reliably understand these stories. 
Despite the long list of supervised methods for story completion and procedural understanding, current AI has no mechanisms to automatically track and explain procedures in unseen stories.
To bridge this gap, we study the ability of AI models to transfer procedural knowledge to novel narrative tasks in a transparent manner. 
We design \method: a comprehensive framework that integrates state-of-the-art modeling architectures, training regimes, and augmentation strategies based on both natural and synthetic stories. To address the lack of densely annotated training data, we devise a robust automatic labeler based on few-shot prompting to enhance the augmented data. 
Our experiments with in- and out-of-domain tasks reveal insights into the interplay of different architectures, training regimes, and augmentation strategies. \method's labeler has a clear positive impact on out-of-domain datasets, while the resulting dense annotation provides native explainability.
\blfootnote{* Work done when KM was at Carnegie Mellon University}
\end{abstract}

\section{Introduction} %(0.75)}

\begin{figure*}[!t]
	\centering
	\includegraphics[width=0.79\textwidth]{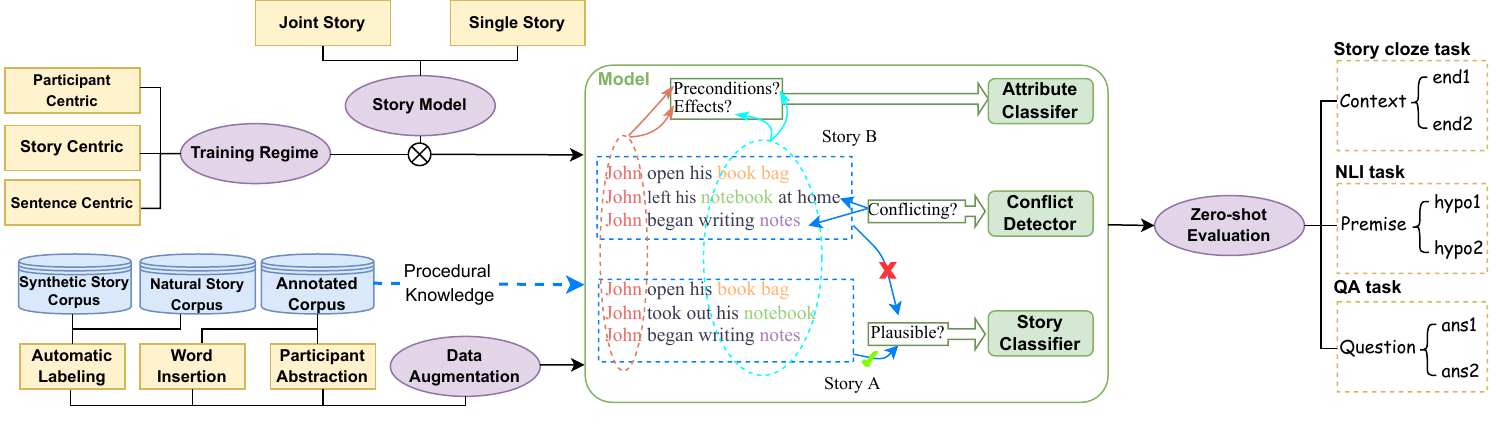}
	\caption{An illustration of the \method~framework, which transfers procedural knowledge from a source task through zero-shot evaluation on unseen tasks. The model structure is presented in the green box.  \method~includes various story modeling strategies and training regimes. To address data sparsity, \method~supports three data augmentation methods: participant abstraction, word insertion, and automatic labeling.}% \filip{Typo: lableing} \yifan{corrected}}
	\label{fig:framework}
\end{figure*}

% Story understanding essential for AI
Building AI agents that understand stories is central to many domains, ranging from cooking~\cite{rajaby-faghihi-kordjamshidi-2021-time} to science~\cite{2018tracking}. 
This is because practically any situation can be associated with a story that requires an agent to judge and explain its plausibility~\cite{charniak1972toward}.
% AI agents that fail to understand an evolving situation are unlikely to be adopted to assist, or collaborate with, people on their everyday tasks.
% Many applications
% Lots of aspects go into understanding stories
% However, trustworthy reasoning over stories is challenging, as it includes reasoning over chains of events, understanding situational frames, and injecting implicit (commonsense) knowledge.
% Example
\textit{How does one decide whether a story is plausible or not?}
Let us consider the two similar stories shown in \autoref{fig:framework}. 
% These factors are essential, for example, to decide that out of the two similar stories in  
Story A makes sense because taking out the notebook is often followed by writing, and a key affordance of notebooks is to enable writing. Meanwhile, story B is implausible, as having the notebook at a different location hinders the possibility of writing in it.
Thus, to understand stories about everyday situations,
% it is intuitive that a person would start writing after they take out their notebook. However, story B is implausible, as a person is unlikely to write notes if they forgot their notebook at home.
% To understand such a situation, 
% What would be ideal
an AI model needs to be able to track the states of the relevant participants, understand the implications of described events, detect anomalous and unexpected behaviors, and project alternative and counterfactual scenarios~\cite{schank1975scripts}.

%% Procedural reasoning is at the core of AI understanding situations
% Building trustworthy AI agents requires a robust and explainable understanding of everyday situations. Agents that fail to understand an evolving situation are unlikely to be adopted to assist, or collaborate with, people on their everyday tasks.  Agents that exhibit these skills in a robust and explainable way may enable AI technologies to be adopted in a wide set of novel applications, ranging from traffic to healthcare.

%% Much work on it over the past years, but it is all supervised, limited evaluation
% \filip{I got lost with this paragraph, not sure what the argument exactly is, let's discuss.}
While story comprehension has been a popular goal over the past decade~\cite{sang2022survey}, state-of-the-art methods typically lack pragmatic inference, and rely heavily on benchmark-specific training, which limits their generalization to novel benchmarks and tasks. 
% Moreover, these methods are designed as black boxes . 
A parallel stream of research \cite{rajaby-faghihi-kordjamshidi-2021-time,gupta-durrett-2019-tracking,ma2022coalescing} has developed methods for state tracking of participants in the domains of science and cooking to increase the model's interpretability. 
A recent task of procedural reasoning about physical processes \cite{storks2021tiered} measures the ability of methods to simultaneously predict the plausible story, the conflicting sentence pairs for the implausible story, and the physical states of the participants.
% state tracking. Given a pair of similar stories, the model should identify (1) the plausible story, (2) conflict sentence pairs in the implausible story, and (3) physical states in sentences bringing conflict. 
Although this task provides a natural bridge between the model's procedural understanding and the overall story assessment, so far it has only been considered in a supervised setting, raising questions about the generalizability of the findings on unseen data.
% by
% an inner view of the model reasoning process and the model with high performance is assumed to understand the procedural knowledge, the credibility is still limited because of its supervised setting. 
% In particular, zero-shot evaluation is an efficient measure of testing model generalisability and verifying whether the model understands procedures. 
To our knowledge, no existing effort has studied the procedural transfer ability of AI models to unseen tasks, nor addressed the inherent lack of densely annotated data about story procedures.
% transfer the understanding of procedural to an unseen task, nor analyzed this understanding through causal reasoning. Moreover, developing models with a transferable understanding of stories and procedures requires densely annotated data, which may not be readily available.

%% This paper: transferring procedural knowledge across tasks
In this paper, we \textit{study the ability of AI models to transfer procedural knowledge across story-based tasks in a transparent manner}.
% the challenge of transferring  in a robust and explainable way. 
Our contributions\footnote{https://github.com/1171-jpg/LEAP} are as follows:
\begin{enumerate}
    \item 
    % A novel zero-shot setting for transferring procedural understanding across tasks. 
   \textbf{A comprehensive framework called \method}~(Learning from Experience by Annotating Procedures), that integrates state-of-the-art language model (LM) architectures, training regimes, and augmentation strategies. \method~is designed to study the ability of models to transfer knowledge about procedures across story tasks. % in a zero-shot setting.
   % , and for study the effect of different design choices on the model generalizability.
    \item \textbf{An automatic labeler} within \method~that densely annotates collected stories based on semantic parsing and few-shot prompting.
    % based on few-shot prompting that reliably annotates procedural narratives assembled through our augmentation strategies. 
    \method's labeler can be generically applied to annotate participants and their attributes in arbitrary realistic and synthetic stories.    
    % Augmentation strategies based on modifications of densely annotated data based on abstraction or word addition, or integration of real or synthetic stories. To aid augmentation, we devise 
    % We apply a few-shot prompting setting in automatic annotation, 
    \item \textbf{An extensive evaluation} of a wide range of models on representative procedural-driven tasks covering different transfer capabilities, task formats, and domains. We provide insights into the strengths and weaknesses of current models to reason over novel stories, and we showcase the ability of the developed labeler to generalize better with existing supervised methods.
\end{enumerate}

% \yifan{We use the word 'story', and 'procedural' a lot in different places on paper. We should make the whole concept consistent.}

\section{Related Work}%(0.5)}
\textbf{Story Understanding} has been a popular task over the past decade, resulting in many popular benchmarks \cite{mostafazadeh-etal-2016-corpus,kocisky-etal-2018-narrativeqa} and methods 
% remove kocisky-etal-2018-narrativeqa
\cite{2019story,Cui_Che_Zhang_Liu_Wang_Hu_2020}.
%remove schwartz-etal-2017-effect
While some previous work has studied the effect of commonsense knowledge on story understanding \cite{2019incorporate}, they only focus on the supervised setting and in-domain evaluation.  % tian-etal-2020-scene
% In \cite{sun2018improving}, language models are enhanced with human reading strategies like highlighting and self-assessment.
% Eventbert~\cite{zhou2022eventbert} is a method that enhances language models with an understanding of the correlation between events. \citeauthor{tian-etal-2020-scene}~[\citeyear{tian-etal-2020-scene}] reasons over story endings by restoring scenes of event sequences based on background commonsense knowledge.
%While these methods integrate interpretable and human-inspired insights, they have been generally evaluated in a supervised setting.
Conversely, our paper focuses on the zero-shot evaluation setting and studies the transfer of procedural knowledge with different data augmentation strategies. %without assuming the availability of benchmark-specific training data. 

\noindent \textbf{Procedural Reasoning} Unlike story understanding task which only requires the final prediction label, procedural reasoning requires the model to track the states of the participants at every step in a process \cite{Bosselut2017SimulatingAD,2018tracking}. Many dedicated methods have been developed for procedural reasoning~\cite{rajaby-faghihi-kordjamshidi-2021-time,gupta-durrett-2019-tracking,zhang2021koala}. 
% DBLP:journals/corr/abs-2003-13878,
More recently, \citeauthor{storks2021tiered} proposed the TRIP benchmark to bridge the gap between procedural reasoning and story understanding, where the model needs to perform tiered reasoning over story pairs. \citeauthor{ma2022coalescing} [\citeyear{ma2022coalescing}] proposed a procedural reasoning model that can be extended to perform story understanding tasks. However, there has been no study of the zero-shot generalization ability of these aforementioned models. 
%jointly predicts participant attributes and their transitions, while TSLM approaches procedural reasoning as a question-answering task with adapted language models enhanced with a timestep encoding~\cite{rajaby-faghihi-kordjamshidi-2021-time}. NCET~\cite{gupta-durrett-2019-tracking} uses a CRF model to model the participant states at each step. KOALA~\cite{zhang2021koala} leverages background commonsense knowledge and optimizes their reasoning with a global method. Similar to our work, CGLI~\cite{ma2022coalescing} combines the task of procedural and story understanding. However, to the best of our knowledge, our framework is the first that investigates story understanding in a zero-shot manner.

\noindent \textbf{In-Context Learning} With the recent progress of large pre-trained LMs, prompting has become a popular approach to tackle many NLP tasks \cite{brown2020language}. More specifically, zero-shot prompting directly feeds in the input to the model to elicit an output \cite{wei2021finetuned}, while few-shot prompting additionally appends example demonstrations to the input to better guide the model's prediction \cite{10.1145/3411763.3451760}. Most of the previous work on prompting studied tasks that require simple outputs, e.g., classification \cite{min2022rethinking}, and only a few have attempted to apply prompting to tasks that require complex structured outputs \cite{madaan2022}.
%remove spiliopoulou2022events
In our work, we not only explore prompting for procedural reasoning but also study the transfer ability of the elicited procedural knowledge to other tasks.

\section{\method: Procedural Transfer Framework} \label{Framework} 

We introduce a framework called \textbf{\method}~(\textbf{L}earning from \textbf{E}xperience by \textbf{A}nnotating \textbf{P}rocedures) that enables the transfer of explicit procedural knowledge from a source task $S$ to target tasks $T_1, T_2, ...$ in a zero-shot manner. \textit{S} consists of pairs of 
plausible and implausible stories $(P, P^{\prime})$. Each story comprises $n$ sentences $s_1,s_2, ... s_n$ and is annotated with three procedural components: (1) the preconditions and the effects for each attribute $a \in A$ of each participant $e \in E$ at every step in the story (or $E^\prime $ for $P^\prime$, which is different from $E$); (2) a pair of conflicting sentences ($s$,$s^\prime$) in the implausible story; and (3) a plausibility label of the story, $P_{plau} \in \{0,1\}$. A model that learned procedural knowledge rather than spurious correlations from the source task should be able to generalize to unseen tasks that require similar kinds of reasoning.
To test this, we select target tasks $T_i$ that are  in multiple-choice format, where given a partial procedure description, a model has to select the answer that optimally completes the procedure.

% \subsection{Core Modeling Architecture}
The model structure, shown in green in the center of \autoref{fig:framework}, is loosely based on CGLI \cite{ma2022coalescing}, which achieves state-of-the-art results on procedural understanding tasks. It takes a pair of stories as inputs to a Transformer language model encoder and then three distinct output layers are used to perform the stratified reasoning. %\textit{Attribute classifier},\textit{Conflict detector}, and \textit{Story classifier}, use the output from the Transformer model to predict 
%three previous procedural \filip{previous? do you mean following?} \yifan{previous refer to the three objects introduced in the first paragraph."The source task consists of ..."}components. 
An \textit{Attribute classifier} predicts the preconditions and the effects for each attribute of each participant. A \textit{Conflict detector} predicts the conflict sentences in the implausible story. A \textit{Story classifier} determines which story is plausible. 
% Building on this general model, 
\method~extends this general model with two story modeling methods, three training regimes, and three augmentation strategies. We also evaluate the \method~variants on four out-of-domain tasks. A comprehensive overview of \method~is shown in \autoref{fig:framework}.
%\yifan{Shall we change the number of augmentation method to 3?}
% \filip{please explain here what is the model architecture like in 1 paragraph. The reader should not be expected to know the other paper}
% \yifan{The detail of each model block is introduced in Separate Strategy, including formula and corresponding notation. And we present Joint Strategy by introducing its additional modification compared to the Separate Strategy. Thus, in here, I just simply introduce each block in the model and its function.}

 \subsection{Story Modeling}
% \km{Is the separate strategy the same as CGLI?}
% \yifan{Basically, they are the same, except we use sigmoid when picking the conflict sentence pair.}
%% input representation of different input strategies, introduce separate strategy first and then introduce joint strategy as an addition modification based on former
%\method~includes two ways of modeling stories: individually and jointly (as pairs). % a separate story modeling strategy and a joint input strategy based on the model decision pattern. 

%\filip{the formalization of this part is good overall, I am just wondering if it becomes a little dense and hard to follow - maybe we can add an example somewhere to help the readability?} \yifan{Maybe we can replace all formulas with plain text and just use formulas or a specific example in input formation and loss training. It is hard for the reader to accept all the complex and detailed ideas in such a short paragraph. Using plain text to introduce at a high level may be better. }

%\noindent\textbf{Single Story Model.} 
Given a story $P$ and a set of participants in the story $E$, we create a separate input sequence based on every participant $e \in E$ following ~\cite{ma2022coalescing}: $C = [CLS] e [ SEP] s_1 [SEP] ... s_n [SEP]$ where $s_1, ... s_n$ are sentences in $P$. We then add timestep embeddings \cite{rajaby-faghihi-kordjamshidi-2021-time} to mark the current reasoning step. The input embeddings and timestep embeddings are summed and encoded by the LM encoder. The [CLS] token representations from the input sequence of every timestep are extracted for output modeling, resulting in $C \in \mathrm{R}^{n \times d}$ where $n$ is the number of steps in the story and $d$ is the hidden dimension.  
%The method of~\cite{ma2022coalescing} models single stories, i.e., it outputs a plausibility score separately for each story $P$. 
%In this setup, we construct a participant-sentence %\filip{what is entity-first?}\yifan{entity-sentence may be more proper.} 
%input representation \cite{gupta-durrett-2019-effective} for each participant at each step of the procedural story. Given an participant $e \in E$, we construct a pseudo question $Q =$ 'where is \{$e$\}' and combine it with n sentences in P, resulting in an input representation $C = [CLS] Q [ SEP] s_1 [SEP] ... s_n [SEP]$. We pass $C$ to a contextual embedding layer to get $C_{emb} \in R^{d\times m}$, where $m$ is the number of tokens and $d$ is the output dimension. Following \cite{rajaby-faghihi-kordjamshidi-2021-time}, we then create a timestamp token (0=padding or pseudo question, 1=past, 2=current, 3=future) for each token in $C_{emb}$, resulting in $T \in R^{m}$, to include step information explicitly. We use another timestamp embedding to get $T_{emb} \in R^{d\times m}$ and pass the sum
%mation \filip{concatenation?} \yifan{We use the summation in our model} 
%of $T_{emb}$ and $C_{emb}$ to the Transformer LM to yield $C_{enc} \in R^{d\times m}$. We concatenate $C_{enc}^{i} \in R^{d}$ of each sentence in P and extract the head token $C_{enc}^{N\prime} \in R^{d\times n}$ to represent the global procedural information, which is passed on the attribute classifier, conflict detector, and story classifier.
%% attribute classifier 
\textit{Attribute classifier} takes the representation of each step $C^{i} \in \mathrm{R}^d$ and passes through two-layer MLP to predict precondition and effect attributes of the current participant. We create a separate classifier for every unique attribute in the task.
%% conflict detector
\textit{Conflict detector} concatenates every pair of step representations [$C^{i}$;$C^{j}$] $\in \mathrm{R}^{2d}$ and processes them through a linear layer. Then, binary classification is applied to every sentence pair to predict whether it has conflicts. We take the mean of the step representations to form a story representation $C_{s} \in \mathrm{R}^d$ for the final label prediction.
% \begin{align}
%     & C_{confl} = Stack(Concat(C_{enc}^{i\prime},C_{enc}^{j \prime})) \\
%     & P_{confl} = Sigmoid(W_{confl} C_{confl})
% \end{align}

\noindent \textbf{Single Story Model} 
 We project $C_s$ to a two-dimensional vector using a linear classifier to represent the plausible and implausible class logits. Thus, each story is classified separately. 
% \filip{similar to CGLI?} \yifan{right, the active function in conflict sentence detector is different.}
% \textit{Story classifier} computes mean of all $C_{enc}^{i\prime}$ and use another linear layer for binary classification. Formally:
%  \begin{align}
%     & C_{plau} = Mean(C_{enc}^{N \prime})\\
%     & P_{plau} = Softmax(W_{plau}^T C_{plau})
% \end{align}
 %\filip{I added a sentence to make the losses more readable, pls validate that it makes sense.} \yifan{The new version is great.}

%% Joint input strategy

\noindent\textbf{Joint Story Model} An obvious drawback of single-story modeling is that the relationship between the stories of the pair is not captured. However, since our model expects a unique input sequence for every participant in the story, it may be hard to construct parallel input sequences for stories that do not have identical participants. To remedy this, we first obtain the common participant set, $E_c = E\cap E^\prime$ and then construct parallel input sequences for every $e \in E_c$ as previously described. Then for unaligned participants, we create a dummy participant $e_{0}$ to fill in the slot in the corresponding input sequence. Hence, both stories will have an equal number of participant-based input sequences. 
In this case, after obtaining $C_s$ and $C_s^\prime$ from the pair of stories, we concatenate these two vectors and perform classification with a linear layer of size $\mathrm{R}^{d \times 1}$ to predict the plausible story. Hence both stories are jointly considered for prediction. 

For both single and joint-story settings, we take the mean of all participant predictions as the final prediction for conflict sentence and story plausibility.% prediction. 
%We jointly optimize over three loss functions corresponding to these classifiers $\textit{L}_{plau}$, $\textit{L}_{confl}$ and $\textit{L}_{attri}$ to train the model with procedural knowledge. $\textit{L}_{attri}$ combines two separate subfunctions: attribute precondition ($\textit{L}_{prec}$) and effect ($\textit{L}_{effe}$).

\subsection{Training Regimes}
%\yifan{check term story-based/story-loss/story focuses through paper.}
As the model optimizes over multiple objectives, we experiment with story, participant, and sentence-centric training regimes.

%% story based
\noindent\textbf{Story Centric} optimizes all three losses, i.e., story, conflict, and attribute loss, for each participant in the story.
% , namely .
% \begin{equation}
%    \textit{L} = \textit{L}_{plau} + \textit{L}_{confl} + \frac{1}{A} \sum_{i=0}^{A} (L_{prec}^{i}+L_{effe}^{i})
% \end{equation}
% Here A is the number of unique attributes, and we set the story label to implausible for all participants from the implausible story.

%% Entity based

\noindent\textbf{Participant Centric} modifies the \textit{story-centric} loss by setting the story label to implausible only for participants that appear in the conflicting sentences of the implausible story. %\filip{does this mean the entities in the conflicting sentence or something more precise?} \yifan{changed the presentation}

%% Sentence based
\noindent\textbf{Sentence Centric} omits the story loss, only optimizing the attribute and conflict losses.
% \begin{equation}
%    \textit{L} = \textit{L}_{confl} + \frac{1}{A} \sum_{i=0}^{A} (L_{prec}^{i}+L_{effe}^{i})
% \end{equation}
To predict the story label in this setting, we obtain the negative sum of all conflicting sentence pair predictions to detect the more plausible story.% is more plausible.

%% augmentation
\subsection{Augmentation Strategies}
Procedural text understanding requires dense annotation of participants and their state attributes
% extraction and rich annotation of their attributes 
in each step. Given the laborious data collection, 
% indicating the laboriousness and challenge of collecting large data. Thus
most existing benchmarks are small in scale, which may hinder the learning of generalizable models. Instead of further manual annotation (e.g., by crowdsourcing), %for more examples, 
we propose a cheap and automatic data augmentation 
% to automatically augment the pr data 
for training procedural reasoning models. 
%For participant extraction, there is no existing work aiming at picking participants directly from the procedural text. For attribute annotation, previous work \cite{zhang2022survey} focuses heavily on active learning to address this issue, which may bring the noise in new data due to the biased distribution of training data \cite{https://doi.org/10.48550/arxiv.2101.11665}. %\filip{this sentence needs citation for active learning and perhaps one for bias in AL}. \yifan{Added} 
%We introduce two approaches for data augmentation: modification of in-domain data and annotation of additional data. 

% introduce three data augmentation strategies, \textit{Automatic Annotation},\textit{Entity Abstraction}, and  \textit{Word Insertion}, to provide a new problem-solving perspective and discuss their effectiveness through comparison. \textit{Automatic Annotation} extract entities and their attribute annotation automatically. \textit{Entity Abstraction} and  \textit{Word Insertion}
 % generate new data from TRIP and inherit attribute annotation from original TRIP directly.

\noindent \textbf{In-Domain Augmentation} We define two in-domain methods: Participant Abstraction and Word Insertion. %, \textit{Participant Abstraction} and \textit{Word Insertion}, which generate new instances from the in-domain training data and inherit the original annotated labels directly. % from original TRIP directly. 
The \textit{Participant Abstraction} augmentation is inspired by \citeauthor{hopner2022leveraging}'s idea to replace specific concepts with more general ones to aid reinforcement learning.
% that replaces specific with generic concepts to aid . 
We replace all non-human participants with their direct hypernym in WordNet~\cite{10.1093/ijl/3.4.235} to generate the new dataset. For example, ``bread" $\rightarrow$ ``baked goods", and ``pan" $\rightarrow$ ``cooking utensil". 
The intuition is that the hypernym shares similar physical properties with the original participant, enabling the reuse of the original attribute annotation. % without bringing the conflict to the whole procedural text.
% \noindent
\textit{Word Insertion}
adds adjectives and adverbs to the existing sentences in the in-domain corpus. It selects suitable words based on a contextual word embedding of the original sentence, obtained with a pre-trained LM, e.g.,
% and updates the original sentence with suitable words. 
% For example, 
``Tom ate the cold soup" $\rightarrow$ ``Tom ate the wonderful cold tomato soup". As word insertion merely enriches the participant information, we directly reuse the attribute annotation.

\noindent \textbf{External Data Augmentation} %The Web contains a vast amount of procedural narratives that can be leveraged as additional data for augmentation. 
We experiment with two kinds of external data: \textit{Natural Stories} and \textit{Synthetic Stories}.
As \textit{Natural Stories}, we select ROCStories~\cite{mostafazadeh-etal-2016-corpus}, a popular story cloze task dataset about everyday events. We use \textit{Synthetic Stories} that are automatically generated from the Commonsense Knowledge Graph (CSKG)~\cite{ilievski2021cskg} based on psychological axioms~\cite{gordon2017formal}. The synthetic dataset, generated by~\cite{ilievski2021story} has over 100k plausible commonsense stories based on three story types with corresponding templates: unmet expectations, substitutions, and object modifications. Each plausible story is associated with an implausible story based on graph patterns. For our experiments, we use a stratified sample of 3K synthetic story pairs.
%A key challenge of the additionally collected stories is that they would typically lack the dense annotation assumed by our explainable framework. 

Unlike in-domain augmentation data, the external stories do not come with the dense attribute annotations assumed by~\method. To bridge this gap, \method~includes a novel labeler that can automatically extract participants and annotate their attributes without training, which we describe next.

% We adapt our labeling system to do automatic annotation on , an artificial QA set created from KG data as well as 
% The detailed component of the labeling system is demonstrated in Section \ref{labeling}.

\section{Automatic Story Labeling} \label{labeling}
% We newly propose a labeling system that can automatically process entity extraction and attribute annotation on any type of data.  \filip{these sentences are moved from 3.3 - should be integrated}

\begin{figure*}[!t]
	\centering
	\includegraphics[width=0.8\textwidth]{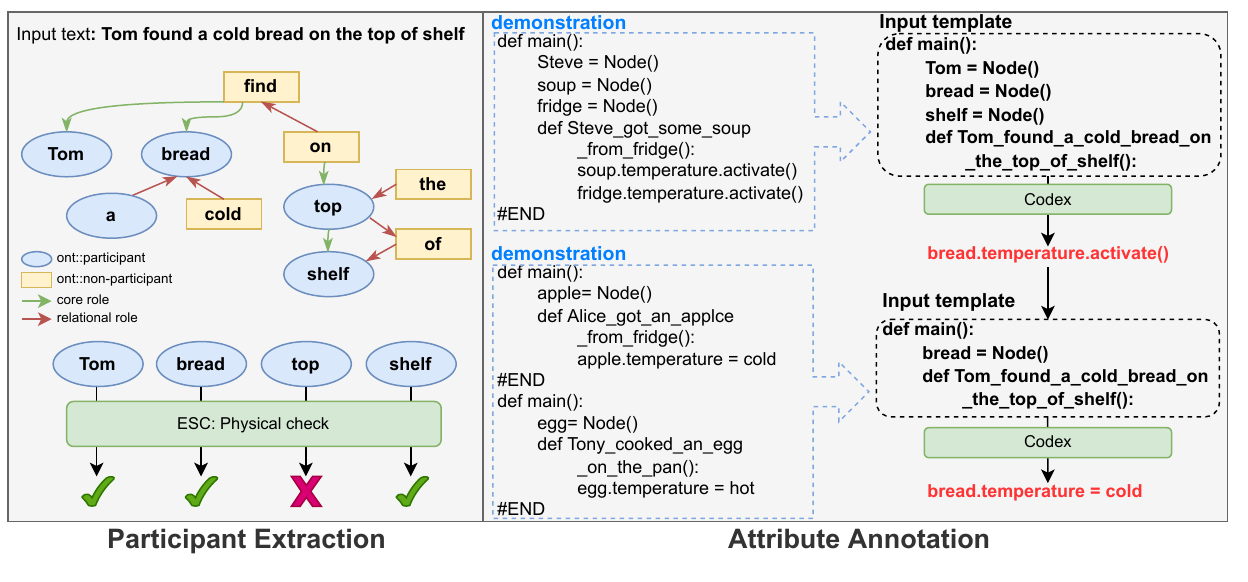}
	\caption{An overview of the \method's labeler. The labeler first extracts all participants in the story. The participants must belong to one of the pre-defined ontology classes, be involved in at least one core role relation, and be physical in nature. For each participant, we construct the corresponding Python function code and adapt few-shot prompting in Codex to label its attribute states automatically.}
	\label{fig:labeling system}
\end{figure*}

%To address the lack of reliable participant extraction and rich attribute state annotation, we propose a novel labeling system that can densely annotate any kind of story. 
\method's labeler consists of \textit{Participant Extraction} and \textit{Attribute Annotation} (Figure \ref {fig:labeling system}). We extract participants based on semantic rules and verify each participant using WordNet. To annotate the attribute states of each participant, we devise a method for code-style few-shot LM prompting. % and employ few-shot prompting with a structure-aware model. %We next describe the design of each part in detail.% in this section.

%% entity extraction
\subsection{Participant Extraction}

\noindent\textbf{Noun Phrase Detection} We first extract nouns and noun phrases from every sentence in the story using

the TRIPS~\cite{Allen2017BroadCD} parser. TRIPS infers a logical form for each sentence and then constructs a tiered ontology that assigns each word to several specific ontology classes based on its semantic information. We manually pick eleven ontology classes (see appendix) and extract all nouns belonging to the classes as possible participants. 
For the noun phrase participants, we   combine participants sharing the same semantic roles in the sentence. This approach allows us to extract composite concepts like ``dog cage'', and even rare phrases like ``guinea pig cage''. 

\noindent\textbf{Core Role Detection} TRIPS divides semantic roles into two groups: core roles indicate a relation between participants, while relational roles indicate lexical item definitions such as adjectives and adverbs. We select nine core roles in which a participant is engaged in a dynamic proposition (see appendix), and only keep the participants involved in core roles. 

\noindent\textbf{Physical Participant Detection} We then filter out non-physical participants because annotated attributes only apply to physical ones. 
To do so, we obtain the WordNet synsets for each participant, and 
% Specifically, for each candidate participant, we query WordNet for its synsets and 
use a word sense disambiguation model
% perform a go through a physical check on all filtered entities and leave all physical entities as the final result of our entity extraction. For each filtered entity, we get all its noun meanings from WordNet and use 
\cite{barba-etal-2021-esc} to pick the appropriate word meaning in the story context. % of a story. %ESC is a Transformer-based model that performs Word Sence Disambiguation by taking input target words with several possible definitions and outputting the most probable input span. 
%After getting the most probable meaning of each candidate participant, 
Finally, we check its physicality by traversing the WordNet hypernym tree of a synset iteratively until reaching the ``physical entity" or ``abstract entity" synset.

%% attribute annotation
\subsection{Attribute Annotation}
%As supervised models may overfit to their source tasks, few-shot prompting unleashes the model's knowledge learned from training data and is getting popular due to its high generalization ability on unseen data~\cite{spiliopoulou2022events}. 
Inspired by recent findings that code-style prompts are more effective for structure-aware tasks \cite{madaan2022}, we propose to convert story inputs into Python-style prompts and leverage Codex~\cite{chen2021evaluating} to generate the attribute annotation. As shown in \autoref{fig:labeling system}, we convert the story prompt into a Python function, where each participant is defined as a Node class and each sentence in the story is converted to a function name. The attribute annotations for a sentence are added as statements in the corresponding function. Then the evaluation example is converted to the same format to elicit outputs. %instead of a natural language generation task can output promising results for  Based on these prior findings, we propose a novel method for converting procedural stories into code format and adapting few-shot prompting with the Codex model~\cite{chen2021evaluating} to automatically label the attribute of each participant. 

A remaining challenge is that our stories have many participants, and we seek to extract a large set of attributes, resulting in large and sparse output space. In other words, only a few attributes for a subset of participants has non-trivial labels at any step of the story, while the rest are irrelevant, which makes the method by \cite{madaan2022} insufficient for our case. To overcome the skewed label distribution, we further decompose the attribute annotation into two steps:
%In particular, ~\cite{min2022rethinking} points out demonstrations in the prompting test should follow the actual distribution. As in most case participants don't have an active attribute label, we design a two-step prompting structure to solve the imbalance distribution issue in label space. In our attribute annotation, we first construct a Python function code for each story to find participants with possible active attributes (\cref{fig:labeling system}, top - right). Then for each found participant, we construct another Python function code based on the participant and the corresponding sentence containing the participant to get the attribute state (\cref{fig:labeling system}, bottom - right).

\noindent\textbf{Active Attribute Detection} As shown in \autoref{fig:labeling system} (top - right), we first prompt Codex to detect the active participants among all possible participants at every step. Here, Codex sees $k$ examples from in-domain data as prompt demonstrations and predicts the active participants for each attribute. 
%In the few-shot prompting setup, the model is provided with several demonstrations and one input template. Each demonstration $d_i$ consists of one input $x_i$ and one label $y_i$, while the input template only contains the input $x^\prime$, which shares the same structure with the inputs in the demonstration. 
%We concatenate $k$ demonstrations with the input template $x^\prime$ to obtain context $C = \{x_1,y_1,x_2,y_2,...,x_k,y_k,x^\prime \}$ and feed $C$ to Codexfor completion.
%\km{We need to explain why it is necessary to decompose the overall process into two steps} \yifan{add explanation at the beginning of the section.}
%For all possible attributes $a \in A$, we perform active attribute detection separately, and our demonstration comes from the in-domain dataset. The input $x_i$ in $d_i$ is the original sentence in the procedural story. The label $y_i$ consists of all the participants with their attributes activated in the sentence. To perform few-shot prompting with Codex, we manually format all demonstrations into a Python function, which is shown in the top right of \cref{labeling}. All participants in the sentence become \textit{Node} objects, and the sentence becomes the \textit{function} name with active participants contained in it. Given a new sentence $x^\prime$ for prompting, we create corresponding \textit{Node} objects and the \textit{function} name and append $x^\prime$ to the demonstrations. Here, Codex is expected to complete the \textit{function} with the active participants. 

\noindent\textbf{Attribute State Annotation} For each active participant %\filip{active participant or attribute?} \yifan{changed the last sentence in the former section. 'Active participants' is a more clear term.}
in the sentence, we perform another round of prompting to label its attribute state, as shown in the bottom right of \autoref{fig:labeling system}. For attribute state annotation, we apply KATE \cite{liu2022makes} to retrieve demonstrations from the in-domain training set. We compute the sentence embedding and the word embedding of the participants for all possible demonstrations and pick the ones with the highest average cosine similarity. Then we apply Codex to generate the attribute state of the participant.

% \filip{emphasize that this way of using Codex is very new}
\section{Experimental Setup}

\noindent \textbf{Datasets} As a source task, we use the recently introduced TRIP~\cite{storks2021tiered} dataset, which contains 800 story pairs in total.
%where the goal is to predict the accuracy of a story, the corresponding conflicting sentence if the story is implausible, and the relevant participant attributes.
Our target tasks are: 1) ROCStories~\cite{mostafazadeh-etal-2016-corpus}, a story cloze task of selecting the plausible story ending out of two options. 
2) PhysicalIQA (PIQA)~\cite{Bisk2020}, a two-choice question-answering dataset where the task is to pick the more plausible continuations out of two. 3) aNLI~\cite{bhagavatula2019abductive}, where given the beginning and the ending of a story, the task is to choose the more plausible abductive hypothesis out of two options. 4) CODAH~\cite{chen2019codah}, a multiple-choice sentence completion dataset where the task is to pick the most commonsensical choice. We select these benchmarks because they represent a variety of tasks that can benefit from procedural reasoning.
We provide further information about the datasets %in Table~\ref{tab:tasks} 
in the appendix.

\begin{table*}[!t]
  \begin{center}
  \small
    \caption{Main evaluation results across four commonsense tasks with different data augmentation methods. We also included the zero-shot evaluation result of previous studies as well as the supervised setting. We reran TRIP and CGLI and obtained their zero-shot evaluation result as baselines. As TRIP, CGLI, and our model follow a participant-based input construction, we ran our labeler on four target tasks to extract participants. We ran our experiments three times with different seeds and reported the average of three runs. On the ROCStories dataset, we only use the CSKG portion of our augmentation data to respect the zero-shot setting - this result is marked with an asterisk (*) in the table. %\filip{look for more supervised baselines} %\filip{supervised LM result probably incorrect}
    %\yifan{Should we introduce the random result?}\filip{yes, though that would be in table 1.}
    }
    % \filip{drop the row of BERT Tian or compute results on test set} \filip{swap OOD datasets, new order: rocstories, codah, piqa, anli.}

    \label{tab:aug}
    \begin{tabular}{l l |r|r r r |r r r r r r} 
     \hline
     \multirow{2}*{\textbf{Training}} & \multirow{2}*{\textbf{Model}} & \textbf{Data}&\multicolumn{3}{c|}{In domain (TRIP)} &  \multicolumn{6}{c}{Out of domain} \\ 
     % \toprule
     \cline{4-11}
    & & \textbf{size} & \bf Acc & \bf Con & \bf Ver  & \bf ROC & \bf CODAH  & \bf PIQA  & \bf aNLI & \bf RICA    \\
    \hline 
      Random & - & - &-  & - & - & 49.5 & 25.1 & 50.2 & 49.6 & 50.7\\
     \hline
     \multirow{2}*{ZSQA} & LM & - & - & - & - & 70.2 & 49.5 & 67.6 & 65.5  & 50.3\\
     & LM w/ CSKG  & 692K & - & - & - & 89.7 & 68.5  & \pmb{72.4} & 70.5 & 51.7\\
     % T5-3b  & - & - & - & 76.7 & 76.6 & - & -\\
     % BERT \cite{tian-etal-2020-scene} & - & - & - & - & - & 91.9 & -\\
      \hline 
         \multirow{4}*{TRIP} & TRIP  & 0.8K & 78.2 & 22.2 & 7.8 & 61.2 & 30.7 & 51.5 & 50.4  & 52.3 \\
         & CGLI  & 0.8K & 94.1 & 77.3 & \textbf{28.0} & 76.9 & 43.3 & 54.4 & 60.0  & 49.6\\
          & \textbf{\method~(no aug.)} & 0.8K & \pmb{97.3} & \pmb{78.4}  & 27.6  & 86.5 & 45.9  & 59.0 & 64.6 & 54.2 \\
         & \textbf{\method~w/ story aug.} & 5.6K  & 97.0 & 70.0 & 11.3 & \textbf{*90.6} & \pmb{68.7} & 68.6 & \pmb{71.8} & \pmb{57.5} \\
    \hline 
             Supervised & LM & - &-  & - & -  & $^\dagger$97.9 & 83.1 & 79.2 & 85.6 & 52.3\\
           %  & HGN \cite{DBLP:journals/corr/abs-2010-12873} & -  & - & - &- &- & 84.3 & - & - \\
    \hline
        
    \end{tabular}
  \end{center}
\end{table*}

\noindent \textbf{Metrics} We evaluate the procedural story understanding performance on the TRIP test dataset with three metrics: 1) \underline{acc}uracy
of story classification, 2) \underline{con}sistency as a proportion of examples where both story and conflict sentences are correctly classified, and 3) \underline{ver}ifiability as a proportion of examples with correct score on the first two metrics and with correct conflicting participant attributes in conflict sentences. 
For zero-shot evaluation, we report accuracy on the corresponding dev sets, as their test sets are not publicly available.

\noindent \textbf{Baselines}
We compare against the following baselines. 1) As zero-shot QA (ZSQA) baselines, we consider the original \textit{RoBERTa-Large} pre-trained model without any adaptation, and \textit{RoBERTa-L (CSKG)} \cite{ma2020knowledgedriven}, which is adapted on 692K synthetic QA pairs generated from CSKG \cite{ilievski2021cskg}. 2) We include baselines  trained on the TRIP data with the same three losses introduced in \autoref{Framework}, namely, the original paper baseline~\cite{storks2021tiered} and \textit{CGLI}~\cite{ma2022coalescing}. 3) To contextualize the results, we also show the random baseline as a lower bound, and the supervised fine-tuned LM as an upper bound of zero-shot evaluation. For ROCstories, the supervised result is on the test dataset as training data is unlabeled and LM is trained on the dev set. We include implementation details in the appendix.% of this paper.

% \noindent \textbf{Hypotheses}
% Based on the understanding of our framework, we propose a set of hypotheses that will be validated in our experiments:
% \yifan{Maybe we should put them in the results part.} \filip{let's skip hypotheses for now.}

\begin{table}[!t]
  \begin{center}
  \small
    \caption{In-domain and out-of-domain zero-shot evaluation results of different \method~models (modeling one$^1$ or jointly two stories$^2$) and training losses (participant-, story-, and sentence-centric). The joint model makes the decision based on both story pairs and cannot be trained with participant-centric loss. Out-of-domain benchmarks: RS = ROCStories, CD = CODAH, QA = PIQA, aN = aNLI, RI = RICA. }

    \label{tab:arch}
    \begin{tabular}{ l | r |r r r r r r r} 
    
     \hline
     \multicolumn{1}{c|}{\textbf{\method}} & \multicolumn{1}{c|}{In domain (TRIP)} &  \multicolumn{4}{c}{Out of domain} \\ 
     % \toprule
     \cline{1-9}
    \ \bf Loss & \bf Acc / \bf Con / \bf Ver & \bf RS & \bf CD& \bf QA  & \bf aN   & \bf RI  \\
      \hline
            $Part^1$  & 94.5 / 70.7 / 21.9 &71.8 & 33.5 & 52.0 & 54.9 & \pmb{54.2}  \\
           $Sto^1$ & 95.8 / 73.2 / 23.2 & 80.6 & 41.4 & 55.0 & 60.5 & 52.1  \\
         $Sent^1$ & 92.9 / 66.3 / 20.9 & 56.0 & 22.9& 46.4 & 49.7  & 49.1  \\ \hline
          $Sto^2$  & \pmb{97.3} / \pmb{78.4} / \pmb{27.6} & \pmb{86.5} & \pmb{45.9}  & \pmb{59.0} & \pmb{64.6} & \pmb{54.2}  \\
          $Sent^2$ & 92.9 / 69.2 / 24.5 & 60.2 & 24.6 & 49.5 & 50.3 & 50.2  \\

    \hline 
    \end{tabular}
  \end{center}
\end{table}

% Table result over the model with best accuracy
\begin{table*}[!th]
  \begin{center}
  \small
    \caption{Comparison of augmentation methods labeled with \method~and CGLI.}% \filip{why is the rocstories different to T1?}}
    \label{tab:albation}
    \begin{tabular}{l | l | r r r |r r r r r r} 
     \hline
     \multicolumn{2}{c|}{\method} & \multicolumn{3}{c|}{In domain (TRIP)} &  \multicolumn{5}{c}{Out of domain} \\ 
     % \toprule
     \cline{1-11}
    \bf Augmentation Data & \bf Labeler & \bf Accuracy & \bf Consistency & \bf Verifiability & \bf ROCStories & \bf CODAH  & \bf PIQA  & \bf aNLI  & \bf RICA\\
      \hline
          No augmentation & - & 97.3 & \pmb{78.4} &  27.6 & 86.5 & 45.9  & 59.0 & 64.6  & 54.2 \\ \hline
           \multirow{3}*{CSKG} & - & 95.7 & 45.1 & 17.4 & 89.6  & 57.9 & 66.0  & 67.4 & 55.5 \\
           & CGLI & 95.9 & 69.1 & 24.5 & 90.0 & 61.6 & 66.1  & 68.1   & 53.2 \\
          & \method & 96.2 & 68.9 & 16.0 &  \pmb{90.6} & 62.1 &  67.2 & 67.9   & 56.3 \\ \hline
          \multirow{3}*{ROCStories} & - & 96.3 & 71.7 & 21.1 & - &  65.1 & 64.6 & 68.4 & 55.3  \\
           & CGLI & 96.7 & 72.8 & 21.7 & - & 61.5 & 63.0 &  68.3  & 53.3 \\
         & \method & 97.0 & 72.7 & 12.0 & - & 62.8 & 64.3 & 68.5  & 56.0 \\ \hline
         \multirow{3}*{CSKG+ROCStories} & -  & 95.3 & 33.6  & 8.9  & - & \pmb{68.7}  & 67.2  & 70.6 & 54.1 \\
          & CGLI  & 96.7 & 70.0  & 27.0 & - &67.7 & 66.5 & 70.0 & 54.6  \\
         & \method  & 97.0 & 70.0 & 11.3 & - & \pmb{68.7} & \pmb{68.6} & \pmb{71.8}  & \pmb{57.5} \\
         \hline
         Participant Abstraction & n/a & \pmb{97.8} & 75.1  & \pmb{30.1} & 85.1  & 43.9 & 57.7  & 63.5 & 54.6    \\
         Word Insertion & n/a  & 97.2  & 74.6  & 26.7 & 83.5  & 41.5 & 57.3  & 60.6  & 53.1  \\

    \hline
    \end{tabular}
  \end{center}
\end{table*}

\section{Results}

Our experiments target six questions: \textit{1) How does \method~perform on in-domain reasoning tasks? 2) How does \method~perform on out-of-domain tasks? 3) What is the optimal \method~architecture for transferring knowledge to out-of-domain stories? 4) What is the effect of various augmentation strategies? 5) How does the \method~labeling method compare to supervised labeling systems? 6) Does the \method~attribute labeling and compositional design help explainability? }

\noindent\textbf{In-Domain Results}
%  shows the results of \method~and relevant baselines on both in- and out-of-domain benchmarks. 
On the in-domain task (\autoref{tab:aug}),
\method~outperforms the procedural understanding baselines that are trained on the same data. Compared to the stronger baseline, CGLI, \method~performs better in terms of accuracy and consistency, and on par in terms of verifiability. We also observe that story augmentation is detrimental to in-domain performance, which we attribute to the distribution shift of the additional data.  

\noindent\textbf{Out-of-Domain Results} The results of the zero-shot transfer experiments (\autoref{tab:aug}) show that \method~outperforms the baselines, demonstrating the strong generalization achieved by model engineering and data augmentation. 
% positive joint impact of story modeling, training regimes, and augmentation. 
Despite using orders of magnitude fewer data for training,
% its much smaller training set, 
\method~still outperforms \textit{RoBERTa-L(CSKG)} on three out of four commonsense story tasks, showing the efficient data utilization of \method.
% even beats the supervised \textit{RoBERTa-L} model on ROCStories. This shows that our model structure has efficient data utilization and also raises doubt about the supervised model's understanding of the story. 
%While we expect that \method~performs well on story cloze tasks, we also note that \method~still performs better than the baselines on the abductive NLI task, which can be attributed to its ability to consider the answers jointly. 
% Even though aNLI is a different inference task that requires abductive reasoning, 
We note that \method's performance is lower than~\cite{ma2020knowledgedriven} on PIQA and only slightly higher on CODAH. While this can be expected given the focus on QA of \cite{ma2020knowledgedriven}, we also hypothesize that the model performance is directly linked to the breadth of required knowledge. We compute the percentage of task participants that are unseen during training with the augmented data, observing that 
% which may be attributed to our small augmentation dataset, as PIQA is generated from the Internet and contains a wider range of participants. Compared to the participants in the original and augmentation dataset, 
66.7\% of participants in PIQA are unseen, compared to 55.0\% for CODAH and 43.7\% for aNLI. 
%\yifan{Based on the final result, this statistic can also be used to explain the struggling performance in CODAH. And CODAH is also a QA task. Should we discuss them together?}
This emphasizes the importance of suitable data augmentation and labeling, which is a key challenge addressed by our paper.
We also note that although the story augmentation hurts the in-domain performance, it improves 4 to 23 absolute points across the datasets for out-of-domain evaluation. This is an interesting phenomenon that we explore further in later analysis. 
% The lack of training data again supports the importance of our labeling system.
% \yifan{One sentence for detail explanation for QA}\yifan{One sentence on supervised setting}

\noindent\textbf{Comparison of \method~Architectures}
% We compare different story modelings and training regimes in Table \ref{tab:arch} and find the joint story model with story-based loss perform best on both in-domain and out-domain.  
Table \ref{tab:arch} shows the impact of story modeling and training regime choices within \method. The model architecture with story pairs and story-centric loss has the best performance on both in- and out-of-domain evaluations. 
% Comparing within the same training regime, 
\textit{Joint} story modeling outperforms \textit{single} story modeling in all metrics, indicating that the model benefits from considering the story pair input together and gaining insight from the direct comparison of stories. Among the different regimes, %\textit{story loss} reaches the best performance for both story modeling strategies. %This suggests that even when the model input is a participant-sentence pair, it still has the ability to extract knowledge from the whole procedural text. 
Since models are able to handle the input globally, \textit{participant-centric loss}, which provides fine-grained and local labels, can bring confusion and lead to a decline in performance. \textit{Sentence-centric} loss has the worst performance for both story modeling options, and its zero-shot evaluation result is similar to or worse than random. \textit{Sentence-centric} loss guides \method~to find conflict sentence pairs in the story, which is reasonable for the in-domain task as each story pair must contain conflicting sentences but does not generalize well to out-of-domain tasks. Our analysis reveals that the incorrect answers on these out-of-domain tasks are often less commonsensical but not necessarily conflicting.
% for the target tasks. In target tasks, the distractors may just be less plausible compared to the correct answer but not guaranteed to be conflict. 
% The model focuses on finding conflict can end up giving a random guess. \yifan{I think the point in this part is obvious and clear. We can focus on other sections of the result.}

\noindent\textbf{Effect of Augmentation}
%\yifan{Show the result of entity extraction in context.}
% In , we observe that 
As data augmentation helps with out-of-domain tasks and is harmful to in-domain tasks (cf. \autoref{tab:aug}), we next investigate the impact of different augmentation strategies and labeling methods on these tasks.
Here we use \method's labeler to extract participants from augmented stories and use either CGLI or \method~ to do annotation because CGLI does not perform participant extraction. The results in \autoref{tab:albation} show that story augmentation leads to lower results in the in-domain setting and improved out-of-domain accuracy regardless of the data partition or labeler, which is consistent with \autoref{tab:aug}. Regarding data splits, CSKG and ROCStories perform comparably to each other as augmentation sources, whereas their combination reaches optimal performance, demonstrating the benefit of combining synthetic and natural data sources. Conversely, augmentation strategies that modify in-domain data increase the model's performance on the in-domain tasks but decrease its performance on the zero-shot evaluation tasks. 
%This is not surprising as these two methods do not bring new knowledge to the model and rather transform the original data, causing the model to overfit to spurious correlations in the in-domain task, e.g., associating story plausibility with sentence structure or specific verbs.
These findings suggest that augmentation with additional data helps \method~to generalize better to unseen stories, whereas modification strategies increase the overfitting to the in-domain data.
% This suggests that models without external story augmentation are overfitting to the in-domain data and cannot generalize well to unseen stories. This also urges the need to evaluate on a diverse set of tasks to truly understand the models' reasoning ability. % which is reasonable considering the noise contained in generated labels. However, the augmentation strategies bring consistent improvement over the baseline without augmentation on the out-of-domain datasets. \yifan{Maybe we can replace the description word 'consistent', it may bring confusion with metric 'consistency'} 

\noindent \textbf{Comparison of Labeling Methods}
%\yifan{The structure is good so far!}
We also compare the labeling methods of CGLI and \method~ in \autoref{tab:albation}. Here, we observe that using CGLI leads to better in-domain performance but \method~ outperforms CGLI on 9 out of 10 %\filip{should be revised once we get all results} 
zero-shot evaluations. This indicates that CGLI fits the TRIP data distribution well, but it may not generalize as well to new tasks. As CGLI uses the TRIP task to learn its attribute extractor, it is possible that it
% One potential problem is that CGLI 
may also learn to fit the dataset artifacts or annotation errors. Meanwhile,
% , which can also inflate the in-domain results, we investigate this through qualitative analysis in the next paragraph. On the other hand, 
\method~leverages few-shot prompting to annotate new stories, which does not directly fit the task and is more generalizable to out-of-domain cases. 
Finally, augmentation without any fine-grained annotations (no labeler setting) leads to a drastic drop in in-domain performance and worse results on out-of-domain when using synthetic stories. This suggests that high-quality fine-grained labels are necessary for achieving robust procedural reasoning. 
%the attribute labels generated by CGLI may strictly overfit the source task and cannot generalize well to unseen data. The observation that models augmented with CGLI obtain the best verifiability 
%but perform even worse than raw data augmentation on 6 zero-shot tasks also supports the suggestion\yifan{One sentence to support the argument - CGLI can be overfitting(maybe a little wordy)}. Overall, the gap between the labeling system and CGLI is a result of supervised overfitting. When adapted to new data, the supervised model can be limited to its training dataset, while our labeling system can benefit from a few-shot prompting setting and show strong generalizability. 

%Training models with more and more raw data will lead to a continuous drop in both consistency and verifiability, indicating models return to black boxes and lose reasoning ability.\yifan{One sentence in the end to show the negative impact of training with raw data.}
%\filip{the next sentences on the intrinsic eval should be decreased but let's return to it after we have the rest of the results in a better shape} %In \cref{tab:albation}, we also verify the effectiveness of the labeling system  in comparison to CGLI.

\noindent \textbf{Intrisic Evaluation of LEAP Labeler} We measure the intrinsic performance of the  two \method's labeler components on TRIP. For \textit{Participant Extraction}, we compare against \textit{spaCy}.\footnote{https://spacy.io/} To extract participants with \textit{spaCy}, we extract nouns from the story and combine them into possible noun phrases based on their location in the text. Compared to the gold participants in each story in TRIP, our labeler reaches 90.0\% precision and 93.5\% recall, while \textit{spaCy} reaches 69.1\% precision and 89.0\% recall. This means that using \textit{spaCy} directly leads to more noisy participant extraction. It also confirms that filtering participants based on core semantic roles and physical properties is more effective. We also compare the attribute annotation of our labeler with TRIP and CGLI on the in-domain task, and observe a gap in favor of the supervised methods, as can be expected. We present the full evaluation for participant and attribute extraction in the appendix. 

\begin{figure}[!t]
	\includegraphics[width=0.5\textwidth]{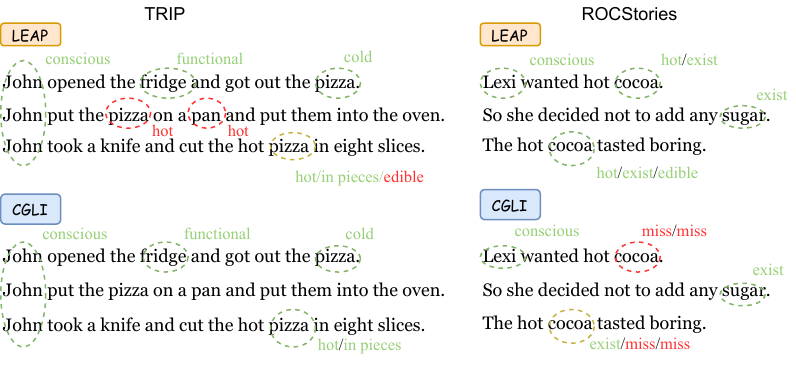}
	\caption{Case study of \method~ and CGLI attribute annotation on in- and out-domain tasks. All active participants are rounded in a dotted cycle with the actual attribute state. Green means correct labeling result, red means incorrect, and yellow means the result contains noise answer.} 
	\label{fig:case}
\end{figure}

\noindent\textbf{Qualitative Analysis}
We conduct qualitative analysis to better understand the attribute labeling of CGLI and \method. \autoref{fig:case} shows the extraction of these two methods on one in-domain (from TRIP) and one out-of-domain (ROCStories) story. 
For TRIP, we use the ground truth attribute annotation to evaluate the result, while for ROCStories, we manually evaluate the labeling result. For the in-domain story (Figure \ref{fig:case}, left), all attribute states produced by CGLI are correct, whereas a small portion of \method's labels is incorrect. %output all correct labels, 
%which is not surprising as CGLI is fine-tuned on TRIP.
% \method~ gives the correct label in most participants while . 
However, we observe that the labels produced by the \method~'s labeler actually make sense even if they do not match the TRIP annotation, e.g., a \textit{hot pizza} is definitely \textit{edible}. This highlights annotation bias in TRIP that can be acquired and amplified by supervised models. On the zero-shot task, we observe that CGLI tends to label fewer participants compared to \method. In particular, in the last sentence, even though CGLI believes that \textit{cocoa} exists, it still misses its \textit{temperature} state. The lower generalizability of CGLI can be attributed to \textit{cocoa} being an unseen participant for CGLI during training.

Besides reaching promising results on zero-shot evaluation, 
\method~is also inherently interpretable through its tiered reasoning process. Consider the following story from ROCStories: \textit{1) I decided to go on a bike ride with my brother. 2) We both headed out in the morning. 3) We were having a lot of fun. 4) Suddenly, he hit a rock and broke his wheel! 5) Watching my brother crash was fun.} For this story, our model infers that the participant \textit{wheel} is not functional

after sentence 4, and that sentence 4 is in conflict with sentence 5. Based on this information, \method~successfully classifies this story as implausible, demonstrating an interpretable and robust reasoning chain. 

% \newpage 
\section{Discussion}%(0.5)}

Our experiments provide insights into the interplay of modeling architectures, training regimes, and augmentation strategies for both in- and out-of-domain reasoning over narratives. Our model, without any augmentation, reaches new state-of-the-art accuracy, consistency, and verifiability on an in-domain task. With augmentation, we are able to generalize much better on out-of-domain tasks, increasing the accuracy by 4 - 23 absolute points across datasets and performing better or competitive to prior baselines. Dense annotation of data either by our novel labeler or by recent work (CGLI) improves the effectiveness of the augmentation data. As our labeler is a few-shot system, it is able to generalize better to unseen stories. Finally, we observe that the joint modeling of stories with a compositional loss function brings the best performance. Our qualitative analysis shows that the better generalization of \method~on out-of-domain tasks is accompanied by robust participant attribute extraction and tiered reasoning.

With these insights in mind, we revisit three assumptions of our work and discuss future directions to improve on them:

% Active learning for selecting stories from a very large pool dynamically
\textbf{Dynamic Selection of Augmentation Stories} While our labeler can in theory be applied to any story, in this paper, we limit the augmentation stories to a single kind of synthetic stories (generated from a CSKG) and a single collection of realistic stories sourced from a popular corpus. Intuitively, the set of stories that can best benefit the model adaptation would depend on the downstream task, e.g., for a commonsense reasoning task, stories with procedural reasoning about household situations may benefit  more than fables or fanfiction stories. While the Web provides an extensive collection of diverse stories, dynamic selection of stories for training augmentation has traditionally been extremely prohibitive due to the high costs of dense annotation. However, our few-shot labeler, enabled by SOTA techniques, opens the possibility for customizable collections of stories to be generated for tasks, or even subtasks. Prior work~\cite{zhangstudy} has investigated sampling 
% remove  swayamdipta2020dataset
methods for a static collection of data; it is a key future work to investigate active learning strategies~\cite{zhang2022survey} for collecting and annotating augmentation data (semi-)automatically.

\textbf{Comprehensive Labeling of Stories} The participant states in this work are described with a relatively rich set of 20 physical attributes, such as temperature and location. As such, our current model is largely geared toward modeling the physical world. Our method can be expanded with complementary aspects of stories, such as the mental states of participants and causal links between events. The psychological axioms by Gordon and Hobbs~\cite{gordon2017formal} and the GLUCOSE dataset~\cite{mostafazadeh2020glucose} with ROCStories event links provide a starting point for both directions, respectively.
Moreover, all of our current attributes, except location, have binary (true/false, or high/low) values as states. Finer-grained annotation, e.g., with qualitative knowledge~\cite{forbus1997qualitative}, can be considered in the future to improve the precision of the reasoning. The combination of manually created theories and resources, on the one hand, with generic LM methods like our labeler, on the other hand, provides an opportunity for a more comprehensive annotation of participant states posed as a generative task.

% Comprehensive labeling of stories (e.g., mental states of entities, causal event links, ...)  - could be turned into a generative setting

\textbf{Comprehensive Understanding of Narratives} Our evaluation in this paper follows the common practice of evaluating on popular benchmarks with short stories, such as ROCStories, CODAH, and TRIP. We believe that the development of methods that understand a wide range of narratives precludes the creation of a diverse set of story benchmarks, including longer stories~\cite{andrus2022enhanced}, fictional stories~\cite{nagarajah2022understanding}, and interactive story-driven games~\cite{wang2022scienceworld}. Curiously, as LM-based reasoning methods have been shown to rely on the surface similarity between the training and the test data (e.g., in terms of token length)~\cite{zhangstudy}, a broadly generalizable method would require the aforementioned dynamic selection of stories, but also novel methods that can abstract over surface properties better.

% Evaluation on more natural stories - scienceworld

\section{Conclusions}

Our work devised \method: a framework for understanding stories through explainable procedural reasoning. \method~consists of modeling architectures, training regimes, and augmentation strategies, selected with the aim to understand both in- and out-of-domain stories. \method~includes a novel labeling method that combines semantic parsing and structure-aware language models to annotate unseen stories in a robust manner. Our experiments showed that joint modeling of stories with a compositional loss function obtained new SOTA results on in-domain tasks. Augmentation with our labeler coupled with external natural or synthetic stories led to a significant increase in performance across out-of-domain tasks, showing strong generalization. In the future, we will build on \method~ to create a method for the dynamic selection of augmentation stories, increase the generality of our labeler to a wider set of attributes with fine-grained state values, and evaluate the result on a representative set of story benchmarks.

\bibliographystyle{named}
\bibliography{ijcai23}

\begin{thebibliography}{}

\bibitem[\protect\citeauthoryear{Allen and Teng}{2017}]{Allen2017BroadCD}
James~F. Allen and Choh~Man Teng.
\newblock Broad coverage, domain-generic deep semantic parsing.
\newblock In {\em AAAI Spring Symposia}, 2017.

\bibitem[\protect\citeauthoryear{Andrus \bgroup \em et al.\egroup
  }{2022}]{andrus2022enhanced}
Berkeley~R Andrus, Yeganeh Nasiri, Shilong Cui, Benjamin Cullen, and Nancy
  Fulda.
\newblock Enhanced story comprehension for large language models through
  dynamic document-based knowledge graphs.
\newblock In {\em Proceedings of the AAAI Conference on Artificial
  Intelligence}, volume~36, pages 10436--10444, 2022.

\bibitem[\protect\citeauthoryear{Barba \bgroup \em et al.\egroup
  }{2021}]{barba-etal-2021-esc}
Edoardo Barba, Tommaso Pasini, and Roberto Navigli.
\newblock {ESC}: Redesigning {WSD} with extractive sense comprehension.
\newblock In {\em Proceedings of the 2021 Conference of the North American
  Chapter of the Association for Computational Linguistics(ACL): Human Language
  Technologies}, pages 4661--4672, Online, June 2021. Association for
  Computational Linguistics.

\bibitem[\protect\citeauthoryear{Bhagavatula \bgroup \em et al.\egroup
  }{2019}]{bhagavatula2019abductive}
Chandra Bhagavatula, Ronan Le~Bras, Chaitanya Malaviya, Keisuke Sakaguchi, Ari
  Holtzman, Hannah Rashkin, Doug Downey, Wen-tau Yih, and Yejin Choi.
\newblock Abductive commonsense reasoning.
\newblock In {\em International Conference on Learning Representations(ICLR)},
  2019.

\bibitem[\protect\citeauthoryear{Bisk \bgroup \em et al.\egroup
  }{2020}]{Bisk2020}
Yonatan Bisk, Rowan Zellers, Ronan~Le Bras, Jianfeng Gao, and Yejin Choi.
\newblock {PIQA: Reasoning about Physical Commonsense in Natural Language}.
\newblock In {\em Thirty-Fourth AAAI Conference on Artificial Intelligence},
  pages 7432--7439, 2020.

\bibitem[\protect\citeauthoryear{Bosselut \bgroup \em et al.\egroup
  }{2018}]{Bosselut2017SimulatingAD}
Antoine Bosselut, Omer Levy, Ari Holtzman, Corin Ennis, Dieter Fox, and Yejin
  Choi.
\newblock Simulating action dynamics with neural process networks.
\newblock In {\em Proceedings of the 6th International Conference for Learning
  Representations (ICLR)}, 2018.

\bibitem[\protect\citeauthoryear{Brown \bgroup \em et al.\egroup
  }{2020}]{brown2020language}
Tom Brown, Benjamin Mann, Nick Ryder, Melanie Subbiah, Jared~D Kaplan, Prafulla
  Dhariwal, Arvind Neelakantan, Pranav Shyam, Girish Sastry, Amanda Askell,
  et~al.
\newblock Language models are few-shot learners.
\newblock {\em Advances in neural information processing systems(NeurIPS)},
  33:1877--1901, 2020.

\bibitem[\protect\citeauthoryear{Charniak}{1972}]{charniak1972toward}
Eugene Charniak.
\newblock {\em Toward a model of children's story comprehension.}
\newblock PhD thesis, Massachusetts Institute of Technology, 1972.

\bibitem[\protect\citeauthoryear{Chen \bgroup \em et al.\egroup
  }{2019a}]{2019incorporate}
Jiaao Chen, Jianshu Chen, and Zhou Yu.
\newblock Incorporating structured commonsense knowledge in story completion.
\newblock {\em Proceedings of the AAAI Conference on Artificial Intelligence},
  33(01):6244–6251, Jul 2019.

\bibitem[\protect\citeauthoryear{Chen \bgroup \em et al.\egroup
  }{2019b}]{chen2019codah}
Michael Chen, Mike D’Arcy, Alisa Liu, Jared Fernandez, and Doug Downey.
\newblock Codah: An adversarially-authored question answering dataset for
  common sense.
\newblock In {\em Proceedings of the 3rd Workshop on Evaluating Vector Space
  Representations for NLP}, pages 63--69, 2019.

\bibitem[\protect\citeauthoryear{Chen \bgroup \em et al.\egroup
  }{2021}]{chen2021evaluating}
Mark Chen, Jerry Tworek, Heewoo Jun, Qiming Yuan, Henrique Ponde de~Oliveira
  Pinto, Jared Kaplan, Harri Edwards, Yuri Burda, Nicholas Joseph, Greg
  Brockman, et~al.
\newblock Evaluating large language models trained on code.
\newblock {\em arXiv preprint arXiv:2107.03374}, 2021.

\bibitem[\protect\citeauthoryear{Cui \bgroup \em et al.\egroup
  }{2020}]{Cui_Che_Zhang_Liu_Wang_Hu_2020}
Yiming Cui, Wanxiang Che, Wei-Nan Zhang, Ting Liu, Shijin Wang, and Guoping Hu.
\newblock Discriminative sentence modeling for story ending prediction.
\newblock {\em Proceedings of the AAAI Conference on Artificial Intelligence},
  34(05):7602--7609, Apr. 2020.

\bibitem[\protect\citeauthoryear{Dalvi \bgroup \em et al.\egroup
  }{2018}]{2018tracking}
Bhavana Dalvi, Lifu Huang, Niket Tandon, Wen-tau Yih, and Peter Clark.
\newblock Tracking state changes in procedural text: a challenge dataset and
  models for process paragraph comprehension.
\newblock {\em Proceedings of the 2018 Conference of the North American Chapter
  of the Association for Computational Linguistics: Human Language
  Technologies, Volume 1 (Long Papers)}, 2018.

\bibitem[\protect\citeauthoryear{Forbus}{1997}]{forbus1997qualitative}
Kenneth~D Forbus.
\newblock Qualitative reasoning., 1997.

\bibitem[\protect\citeauthoryear{Gordon and Hobbs}{2017}]{gordon2017formal}
Andrew~S Gordon and Jerry~R Hobbs.
\newblock {\em A formal theory of commonsense psychology: How people think
  people think}.
\newblock Cambridge University Press, 2017.

\bibitem[\protect\citeauthoryear{Gupta and
  Durrett}{2019}]{gupta-durrett-2019-tracking}
Aditya Gupta and Greg Durrett.
\newblock Tracking discrete and continuous entity state for process
  understanding.
\newblock In {\em Proceedings of the Third Workshop on Structured Prediction
  for {NLP}}, pages 7--12, Minneapolis, Minnesota, June 2019. Association for
  Computational Linguistics.

\bibitem[\protect\citeauthoryear{H{\"o}pner \bgroup \em et al.\egroup
  }{2022}]{hopner2022leveraging}
Niklas H{\"o}pner, Ilaria Tiddi, and Herke van Hoof.
\newblock Leveraging class abstraction for commonsense reinforcement learning
  via residual policy gradient methods.
\newblock In {\em 31st International Joint Conference on Artificial
  Intelligence, IJCAI 2022}, pages 3050--3056. International Joint Conferences
  on Artificial Intelligence Organization, 2022.

\bibitem[\protect\citeauthoryear{Ilievski \bgroup \em et al.\egroup
  }{2021a}]{ilievski2021story}
Filip Ilievski, Jay Pujara, and Hanzhi Zhang.
\newblock Story generation with commonsense knowledge graphs and axioms.
\newblock In {\em Workshop on Commonsense Reasoning and Knowledge Bases}, 2021.

\bibitem[\protect\citeauthoryear{Ilievski \bgroup \em et al.\egroup
  }{2021b}]{ilievski2021cskg}
Filip Ilievski, Pedro Szekely, and Bin Zhang.
\newblock Cskg: The commonsense knowledge graph.
\newblock In {\em European Semantic Web Conference}, pages 680--696. Springer,
  2021.

\bibitem[\protect\citeauthoryear{Ko{\v{c}}isk{\'y} \bgroup \em et al.\egroup
  }{2018}]{kocisky-etal-2018-narrativeqa}
Tom{\'a}{\v{s}} Ko{\v{c}}isk{\'y}, Jonathan Schwarz, Phil Blunsom, Chris Dyer,
  Karl~Moritz Hermann, G{\'a}bor Melis, and Edward Grefenstette.
\newblock The {N}arrative{QA} reading comprehension challenge.
\newblock {\em Transactions of the Association for Computational
  Linguistics(TACL)}, 6:317--328, 2018.

\bibitem[\protect\citeauthoryear{Li \bgroup \em et al.\egroup
  }{2019}]{2019story}
Zhongyang Li, Xiao Ding, and Ting Liu.
\newblock Story ending prediction by transferable bert.
\newblock {\em Proceedings of the Twenty-Eighth International Joint Conference
  on Artificial Intelligence(IJCAI)}, Aug 2019.

\bibitem[\protect\citeauthoryear{Liu \bgroup \em et al.\egroup
  }{2022}]{liu2022makes}
Jiachang Liu, Dinghan Shen, Yizhe Zhang, Bill Dolan, Lawrence Carin, and Weizhu
  Chen.
\newblock What makes good in-context examples for gpt-3?
\newblock {\em DeeLIO 2022}, page 100, 2022.

\bibitem[\protect\citeauthoryear{Ma \bgroup \em et al.\egroup
  }{2021}]{ma2020knowledgedriven}
Kaixin Ma, Filip Ilievski, Jonathan Francis, Yonatan Bisk, Eric Nyberg, and
  Alessandro Oltramari.
\newblock Knowledge-driven data construction for zero-shot evaluation in
  commonsense question answering.
\newblock In {\em Proceedings of the AAAI Conference on Artificial
  Intelligence}, volume~35, pages 13507--13515, 2021.

\bibitem[\protect\citeauthoryear{Ma \bgroup \em et al.\egroup
  }{2022}]{ma2022coalescing}
Kaixin Ma, Filip Ilievski, Jonathan Francis, Eric Nyberg, and Alessandro
  Oltramari.
\newblock Coalescing global and local information for procedural text
  understanding.
\newblock In {\em Proceedings of the 29th International Conference on
  Computational Linguistics}, pages 1534--1545, 2022.

\bibitem[\protect\citeauthoryear{Madaan \bgroup \em et al.\egroup
  }{2022}]{madaan2022}
Aman Madaan, Shuyan Zhou, Uri Alon, Yiming Yang, and Graham Neubig.
\newblock Language models of code are few-shot commonsense learners.
\newblock In {\em Conference on Empirical Methods in Natural Language
  Processing (EMNLP)}, Abu Dhabi, UAE, December 2022.

\bibitem[\protect\citeauthoryear{Miller \bgroup \em et al.\egroup
  }{1990}]{10.1093/ijl/3.4.235}
George~A. Miller, Richard Beckwith, Christiane Fellbaum, Derek Gross, and
  Katherine~J. Miller.
\newblock {Introduction to WordNet: An On-line Lexical Database*}.
\newblock {\em International Journal of Lexicography}, 3(4):235--244, 12 1990.

\bibitem[\protect\citeauthoryear{Min \bgroup \em et al.\egroup
  }{2022}]{min2022rethinking}
Sewon Min, Xinxi Lyu, Ari Holtzman, Mikel Artetxe, Mike Lewis, Hannaneh
  Hajishirzi, and Luke Zettlemoyer.
\newblock Rethinking the role of demonstrations: What makes in-context learning
  work?
\newblock In {\em EMNLP}, 2022.

\bibitem[\protect\citeauthoryear{Mostafazadeh \bgroup \em et al.\egroup
  }{2016}]{mostafazadeh-etal-2016-corpus}
Nasrin Mostafazadeh, Nathanael Chambers, Xiaodong He, Devi Parikh, Dhruv Batra,
  Lucy Vanderwende, Pushmeet Kohli, and James Allen.
\newblock A corpus and cloze evaluation for deeper understanding of commonsense
  stories.
\newblock In {\em Proceedings of the 2016 Conference of the North {A}merican
  Chapter of the Association for Computational Linguistics: Human Language
  Technologies}, pages 839--849, San Diego, California, June 2016. Association
  for Computational Linguistics.

\bibitem[\protect\citeauthoryear{Mostafazadeh \bgroup \em et al.\egroup
  }{2020}]{mostafazadeh2020glucose}
Nasrin Mostafazadeh, Aditya Kalyanpur, Lori Moon, David Buchanan, Lauren
  Berkowitz, Or~Biran, and Jennifer Chu-Carroll.
\newblock Glucose: Generalized and contextualized story explanations.
\newblock In {\em Proceedings of the 2020 Conference on Empirical Methods in
  Natural Language Processing (EMNLP)}, pages 4569--4586, 2020.

\bibitem[\protect\citeauthoryear{Nagarajah \bgroup \em et al.\egroup
  }{2022}]{nagarajah2022understanding}
Thiloshon Nagarajah, Filip Ilievski, and Jay Pujara.
\newblock Understanding narratives through dimensions of analogy.
\newblock In {\em Workshop on Qualitative Reasoning (QR)}, 2022.

\bibitem[\protect\citeauthoryear{Rajaby~Faghihi and
  Kordjamshidi}{2021}]{rajaby-faghihi-kordjamshidi-2021-time}
Hossein Rajaby~Faghihi and Parisa Kordjamshidi.
\newblock Time-stamped language model: Teaching language models to understand
  the flow of events.
\newblock In {\em Proceedings of the 2021 Conference of the North American
  Chapter of the Association for Computational Linguistics(ACL): Human Language
  Technologies}, pages 4560--4570, Online, June 2021. Association for
  Computational Linguistics.

\bibitem[\protect\citeauthoryear{Reynolds and
  McDonell}{2021}]{10.1145/3411763.3451760}
Laria Reynolds and Kyle McDonell.
\newblock Prompt programming for large language models: Beyond the few-shot
  paradigm.
\newblock In {\em Extended Abstracts of the 2021 CHI Conference on Human
  Factors in Computing Systems}, CHI EA '21, New York, NY, USA, 2021.
  Association for Computing Machinery.

\bibitem[\protect\citeauthoryear{Sang \bgroup \em et al.\egroup
  }{2022}]{sang2022survey}
Yisi Sang, Xiangyang Mou, Jing Li, Jeffrey Stanton, and Mo~Yu.
\newblock A survey of machine narrative reading comprehension assessments.
\newblock In {\em 31st International Joint Conference on Artificial
  Intelligence, IJCAI 2022}, pages 5580--5587. International Joint Conferences
  on Artificial Intelligence, 2022.

\bibitem[\protect\citeauthoryear{Schank and Abelson}{1975}]{schank1975scripts}
Roger~C Schank and Robert~P Abelson.
\newblock Scripts, plans, and knowledge.
\newblock In {\em IJCAI}, volume~75, pages 151--157, 1975.

\bibitem[\protect\citeauthoryear{Storks \bgroup \em et al.\egroup
  }{2021}]{storks2021tiered}
Shane Storks, Qiaozi Gao, Yichi Zhang, and Joyce Chai.
\newblock Tiered reasoning for intuitive physics: Toward verifiable commonsense
  language understanding.
\newblock In {\em Findings of Conference on Empirical Methods in Natural
  Language Processing (EMNLP) 2021}, 2021.

\bibitem[\protect\citeauthoryear{Wang \bgroup \em et al.\egroup
  }{2022}]{wang2022scienceworld}
Ruoyao Wang, Peter~Alexander Jansen, Marc-Alexandre C{\^o}t{\'e}, and
  Prithviraj Ammanabrolu.
\newblock Scienceworld: Is your agent smarter than a 5th grader?
\newblock In {\em Conference on Empirical Methods in Natural Language
  Processing}, 2022.

\bibitem[\protect\citeauthoryear{Wei \bgroup \em et al.\egroup
  }{}]{wei2021finetuned}
Jason Wei, Maarten Bosma, Vincent Zhao, Kelvin Guu, Adams~Wei Yu, Brian Lester,
  Nan Du, Andrew~M Dai, and Quoc~V Le.
\newblock Finetuned language models are zero-shot learners.
\newblock In {\em International Conference on Learning Representations}.

\bibitem[\protect\citeauthoryear{Zhang \bgroup \em et al.\egroup
  }{2021}]{zhang2021koala}
Zhihan Zhang, Xiubo Geng, Tao Qin, Yunfang Wu, and Daxin Jiang.
\newblock Knowledge-aware procedural text understanding with multi-stage
  training.
\newblock In {\em {WWW} '21: The Web Conference 2021, Ljubljana, Slovenia,
  April 19--23, 2021}, 2021.

\bibitem[\protect\citeauthoryear{Zhang \bgroup \em et al.\egroup
  }{2022a}]{zhangstudy}
Jiarui Zhang, Filip Ilievski, Kaixin Ma, Jonathan Francis, and Alessandro
  Oltramari.
\newblock A study of zero-shot adaptation with commonsense knowledge.
\newblock {\em Automated Knowledge Base Construction(AKBC)}, 2022.

\bibitem[\protect\citeauthoryear{Zhang \bgroup \em et al.\egroup
  }{2022b}]{zhang2022survey}
Zhisong Zhang, Emma Strubell, and Eduard Hovy.
\newblock A survey of active learning for natural language processing.
\newblock In {\em Proceedings of the 2022 Conference on Empirical Methods in
  Natural Language Processing}, pages 6166--6190, Abu Dhabi, United Arab
  Emirates, December 2022. Association for Computational Linguistics.

\end{thebibliography}
e\clearpage

\appendix

\begin{table*}[t!]
  \begin{center}
    \caption{
    Overview of tasks with properties.    }
    \label{tab:tasks}
    \begin{tabular}{c c c c c c} 
     \toprule
    \bf Setup/Dim & \bf Transfer & \bf Task format & \bf Domain & \bf \# Sentences & \bf Style \\
      \midrule
       \bf TRIP & in-domain & story-to-story & physical & three & narrative  \\
       \bf ROCStories & zero-shot & story-to-story & social & five & narrative \\
       \bf CODAH & zero-shot & story-to-QA& commonsense & dynamic  & narrative \\
       \bf PhysicalIQA & zero-shot & story-to-QA & physical & flexible & imperative \\
       \bf aNLI & zero-shot & story-to-NLI & social & three & narrative \\
    \bottomrule
    \end{tabular}
  \end{center}
\end{table*}

\begin{table*}[!h]
  \begin{center}
  \small
    \caption{Ablation results of removing different loss components.}% \filip{why is the rocstories different to T1?}}
    \label{tab:loss}
    \begin{tabular}{l | c c c |c c c c c } 
     \hline
        & \multicolumn{3}{c|}{In domain (TRIP)} &  \multicolumn{4}{c}{Out of domain} \\ 
     % \toprule
     \cline{1-9}
    \bf Ablation & \bf Accuracy & \bf Consistency & \bf Verifiability & \bf ROCStories & \bf CODAH  & \bf PIQA  & \bf aNLI    \\
      \hline
         \method & \pmb{97.3} & \pmb{78.4} &  \pmb{27.6} & \pmb{86.5} & 45.9  & 59.0 & 64.6   \\
        no conflict sentence loss & 97.1 & 8.7  & 2.6  & 86.1  & 46.9 & \pmb{59.3}  & \pmb{65.1}     \\
        no entity attribute loss  & 94.8 & 21.2  & 0  & 77.8  & \pmb{47.4} & 58.4  & 61.8     \\   
        no conflict sentence \& entity attribute loss & 94.6 & 12.4 & 0 & 79.6 & 43.3 & 57.0 & 61.3   \\  
    \hline
    \end{tabular}
  \end{center}
\end{table*}

\section*{Appendix}
%\yifan{I forget some of my notes in ISI, I will make them up tomorrow}
\subsection*{Summary of Datasets}
Table \ref{tab:tasks} shows the summary information of all datasets. 
TRIP differs from the other tasks, as it is the only benchmark for which the system has access to the training data. In terms of the format, TRIP and ROCStories are story cloze tasks, PhysicalIQA and  CODAH are multiple-choice question answering, aNLI is a natural language inference task. Like TRIP, PhysicalIQA concerns the physical reasoning domain, while ROCStories and aNLI are predominantly about social reasoning. CODAH focuses on commonsense knowledge, which is easy for humans but challenging for AI models. Finally, the length of TRIP and ROCStories is five sentences, aNLI has three, PhysicalIQA and CODAH have a dynamic set of sentences. PIQA is written in imperative language, while the other benchmarks use a narrative style.

\subsection*{Ontology Classes and Core Roles Used for Participant Extraction}
We choose the following eleven main ontology classes as the candidate for our participant classes:
(1)\textit{ONT::THE}: a definite singular form. (2)\textit{ONT::THE-SET}: definite PLURAL form. (3)\textit{ONT::A}: an indefinite singular form. (4)\textit{ONT::INDEF-SET}: indefinite plural. (5)\textit{ONT::SM}: Indefinite mass-term(some quantity of). (6)\textit{ONT::PRO}: a pronoun form. (7)\textit{ONT::PRO-SET}: a plural pronoun. (8)\textit{ONT::BARE}: ambiguous noun phrases. (9)\textit{ONT::QUANTIFIER}: universally quantified constructions. (10)\textit{ONT::WH-TERM}: 'wh' terms in questions (which truck). (11)\textit{ONT::WH-TERM-SET}: plural 'wh' terms in questions (which trucks).

Most ontology classes follow general definitions, and we need more information to resolve the actual participants from context. Thus, we analyze each participant with the following nine core roles: (1)\textit{role::AGENT}: the participant plays a role in causing change. (2)\textit{role::AFFECTED}: the participant is changed in an event. (3)\textit{role::FIGURE}: the participant is characterized with other objects. (4)\textit{role::GROUND}: the participant is used to characterize other objects. (5)\textit{role::AFFECTED-RESULT}: the participant undergoes a change during the event(special case for AFFECTED). (6)\textit{role::BENEFICIARY}: the participant benefits from an action.
(7)\textit{role::NEUTRAL}: the participant has existence in an event. (8)\textit{role::EXPERIENCER}: the participant has existence in a cognition and perception event(special case for NEUTRAL). (9)\textit{role::FORMAL}: the participant is involved in a causal argument.

\subsection*{Implementation and Model Details}

For \textit{RoBERTa-L}, we load the weights and parameters from the Transformers library\footnote{https://github.com/huggingface/transformers} and use it to
 score the QA pair directly. For \textit{RoBERTa-L (CSKG)}, \textit{TRIP}, and \textit{CGLI}, we download the published best model and adjust the target task format to fit its evaluation procedure. %Rthe evaluation method in the paper. 
All our models are fine-tuned on the TRIP dataset with or without data augmentations for 15 epochs. We pick the model with the best accuracy on the TRIP test dataset to process zero-shot evaluation. In particular, adapting \textit{TRIP}, \textit{CGLI}, and our models requires knowing the participants in stories. We use the participant extraction result of our \method~'s labeler to provide participants in all target tasks. The Word Insertion augmentation is based on the nlpaug library.\footnote{https://nlpaug.readthedocs.io/en/latest/index.html\#}
%The hyperparameters and computing resources are shown in the appendix. %\yifan{appendix.}

% \subsection*{Model training} 
In all training processes in experiments, we used learning rate $1\mathrm{e}^-6$, batch size 1, gradient accumulation 2, and optimizer AdamW with 0 warm-up steps. We adjust the weights of loss to ensure the summation equals 1. Models trained with \textbf{Story-Centric loss} and \textbf{Participant-Centric loss} have weight 0.4, 0.4, 0.1, 0.1 for $\textit{L}_{prec}$, $\textit{L}_{effe}$, $\textit{L}_{confl}$, $\textit{L}_{plau}$. Models trained with \textbf{Sentence-Centric loss} have weight 0.4, 0.4, 0.2 for $\textit{L}_{prec}$, $\textit{L}_{effe}$, $\textit{L}_{confl}$.
For augmentation methods without \method~'s labeler or CGLI, we only assign 1 for $\textit{L}_{plau}$.
\subsection*{Computing Resources}
For GPUs, we run our experiments on Nvidia RTX A5000. For libraries, we use Pytorch 1.13.0, transformers 4.12.2, and sentence-transformer 2.2.2. For few-shot prompting on Codex, we use code-DaVinci-002 with 0 temperature.

\subsection*{Additional Results}
Table~\ref{tab:loss} shows the results of \method~with different loss ablations.
 Table \ref{tab:step-wise} shows each step's precision and recall in Participant Extraction.  \autoref{tab:participants} shows the statistic analysis of the participants overlapping between target tasks and training data. An evaluation of the different labelers is provided in \autoref{tab:label}.

\begin{table}[!h]
  \begin{center}
    \caption{Step-wise evaluation result of Participant Extraction. NP, CR, and PP are short for Noun Phrase Detection, Core Role Detection, and Physical Participant Detection, respectively.}  
    \label{tab:step-wise}
    \small
    \begin{tabular}{l l c c} 
     \toprule
    \bf Step & \bf  Precision & \bf Recall \\
      \midrule
       NP & 78.2 & 97.9 \\
       NP+CR& 79.3 & 97.0 \\
       NP+CR+PP & 90.0 & 93.5   \\
      \midrule
      \textit{spaCy} & 69.1 & 89.0   \\
    \bottomrule
    \end{tabular}
  \end{center}
\end{table}

% \subsection*{Participants overlapping}

% Labeling System result
\begin{table}[!h]
  \begin{center}
    \caption{Participants overlapping result for each target task.}
    
    \label{tab:participants}
    \small
    \begin{tabular}{l l c c} 
     \toprule
    \bf Target task & \bf  Num of Parti & \bf Unseen Parti. \\
      \midrule
       CODAH & 3899 & 2146(55.0\%) \\
       PIQA& 3994 & \pmb{2663(66.7\%)}  \\
       aNLI & 2134 &932(43.7\%)   \\
    \bottomrule
    \end{tabular}
  \end{center}
\end{table}

% \subsection*{Labeler evaluation}

% Labeling System result
\begin{table}[!h]
  \begin{center}
    \caption{Labeling result on attribute states annotation. }

    % \km{We should be clear that the first two rows are supervised, and it's probably better to separate the categories of methods (supervised vs few-shot)}
    
    \label{tab:label}
    \small
    \begin{tabular}{l l c c} 
     \toprule
    \bf Labeler & \bf Type & \bf Prec.f1 & \bf Eff.f1  \\
      \midrule
       TRIP& supervised &51.2 & 51.2  \\
       CGLI & supervised&72.1 & 75.6  \\
       \method & few-shot &61.2 & 70.3  \\
    \bottomrule
    \end{tabular}
  \end{center}
\end{table}

% \subsection*{Labeler generalizability}
We take a further statistic over all labels generated by both models on ROCStories.
We show the detailed effect and precondition result in Table \ref{tab: participants on ROC}.
 \method~'s labeler will generate labels for nearly three times more participants than CGLI in the average of all attributes, and also participants CGLI generated labels share 32.2\% with TRIP while \method~'s labeler only shares 25.6\%.   We do not consider human attributes here, as overlapping or comparison between human names is meaningless.

%  \yifan{Use some case studies to strengthen this point.}
% \yifan{We need to explain why ROCStories without labeling perform better.}
% \yifan{Introduce another annotation to show the local best result within each category (CSKG, ROCStories)?}
% \yifan{The ablation study on loss is a little weird here, this part can be used to support our discussion in section 2. 1) Conflict Sentence information is not so important.2) Entity attribute information is important for the model to understand the story.}

\begin{table}[!h]
  \begin{center}
    \caption{A statistic over participants \method~'s labeler and CGLI labeled for precondition and effect attribute on ROCStories. The first two rows show the participants overlapping with the source task. The last two rows show the participants' number ratio between the two labelers. All the results are averaged over 15 non-human attributes.}

    \label{tab: participants on ROC}
    \small
    \begin{tabular}{l | l | c c|c} 
     \hline
    \bf Staticitc & \bf Labeler & \bf Prec. & \bf Eff. & \textbf{Average} \\
      \hline
      \multirow{2}*{Participants Overlapping} 
       & CGLI & 29.7\%  & 34.7\%  & 32.2\%\\
       & \method & 25.8\%   & 25.3\%   & 25.6\%\\
    \hline
       \multirow{2}*{Participants Ratio}
              & CGLI & 3.21   & 2.73 & 2.97 \\
             & \method & 1   & 1  & 1 \\
       \hline
    \end{tabular}
  \end{center}
\end{table}

% \subsection*{Loss ablations}

% Table result over the model with best accuracy

\end{document}